%% file: main.tex
\documentclass[11pt,a4paper]{article}
\usepackage[hyperref]{emnlp2020}
\usepackage{times}
\usepackage{latexsym}

\usepackage{booktabs} 
\usepackage{multirow} 
\usepackage{graphicx}
\usepackage{subcaption}
\usepackage{amsmath}
\usepackage{mathptmx}
\usepackage{color}
\usepackage{dirtytalk}

\usepackage{microtype}

\aclfinalcopy 

\title{Multi-stream BERT Representations For Video Question Answering}
\title{MMFT-BERT: Multimodal Fusion Transformer with BERT Encodings for Visual Question Answering}

\author{Aisha Urooj Khan 
  \qquad
  Amir Mazaheri 
 \qquad
  Niels Da Vitoria Lobo 
  \qquad
  Mubarak Shah \\
  Center for Research in Computer Vision, University of Central Florida\\
   \\
  \texttt{aishaurooj@gmail.com, amirmazaheri@knights.ucf.edu}\\ \texttt{shah@crcv.ucf.edu, niels@cs.ucf.edu}
}

\begin{document}
\maketitle
\input{Abstract}

\input{Intro}


\input{Related} 
\input{main_table}
\input{Approach}


\input{test-public}

\input{TablesAndFigures}

\input{Experimental}


\input{Conclusion}

\section*{Acknowledgments}
We thank the reviewers for their helpful feedback. This research is supported by the Army Research Office under Grant Number W911NF-19-1-0356. The views and conclusions contained in this document are those of the authors and should not be interpreted as representing the official policies, either expressed or implied, of the Army Research Office or the U.S. Government. The U.S. Government is authorized to reproduce and distribute reprints for Government purposes notwithstanding any copyright notation herein.

\bibliographystyle{acl_natbib}
\bibliography{emnlp2020}


\appendix
\section{Supplementary Material}

\input{supplementary}

\end{document}

%% file: Abstract.tex
\begin{abstract}

We present MMFT-BERT (\textbf{M}ulti\textbf{M}odal \textbf{F}usion \textbf{T}ransformer with \textbf{BERT} encodings), to solve Visual Question Answering (VQA) ensuring individual and combined processing of multiple input modalities. Our approach benefits from processing multimodal data (video and text) adopting the BERT encodings individually and using a novel transformer-based fusion method to fuse them together.
Our method decomposes the different sources of modalities, into different BERT instances with similar architectures, but variable weights. 
This  achieves SOTA results on the TVQA dataset. 
Additionally, we provide TVQA-Visual, an isolated diagnostic subset of TVQA, which strictly requires the knowledge of visual (V) modality based on a human annotator's judgment. This set of questions helps us to study the model's behavior and the challenges TVQA poses to prevent the achievement of super human performance. Extensive experiments show the effectiveness and superiority of our method\footnote{Code will be available at \url{https://github.com/aurooj/MMFT-BERT}}.
\end{abstract}


%% file: Intro.tex
\begin{figure}[t]
\begin{center}
 \includegraphics[width=\linewidth]{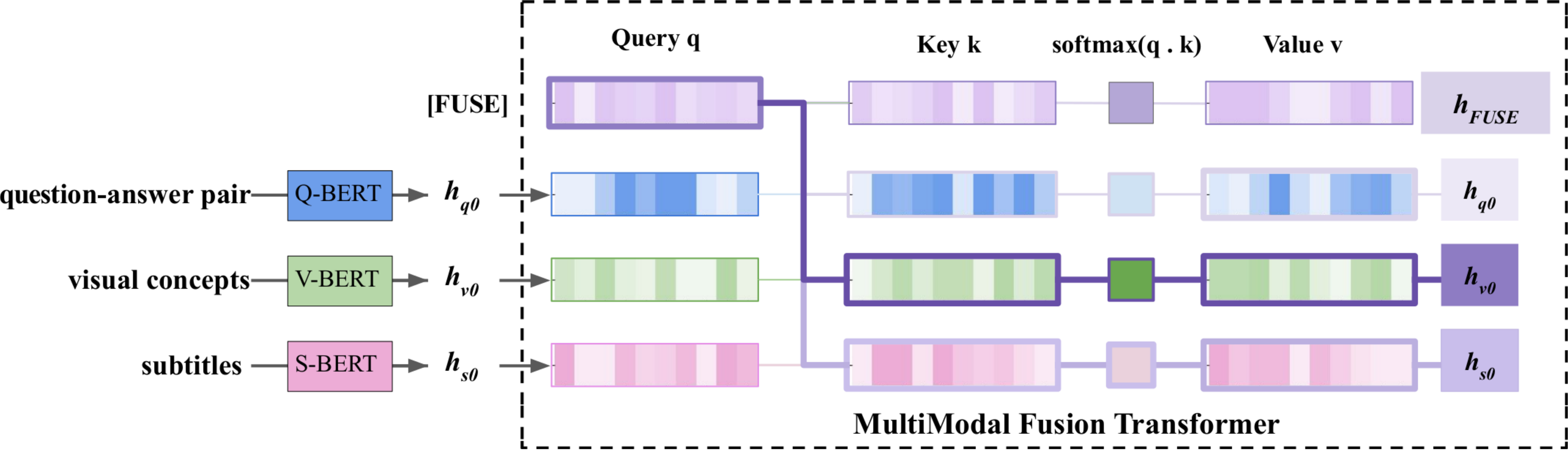}
 \vspace{-15pt}
\end{center}
  \caption{MultiModal Fusion Transformer (MMFT): 
  We treat input modalities as a sequence. [FUSE] is a trainable vector; $ h_{{q_{0}}_j}, h_{{v_{0}}_j}$, and $h_{{s_{0}}_j}$ are fixed-length features aggregated over question-answer (QA) pairs, visual concepts, and subtitles. Using a transformer encoder block, [FUSE] attends all source vectors and assigns weights based on the importance of each input source. Training end to end for VQA enables the MMFT module to learn to aggregate input sources w.r.t. the nature of the question. For illustration purposes, we show that for a single head, MMFT collects more knowledge from the visual source $h_{{v_{0}}_j}$ (green colored) than from the QA and subtitles. Best viewed in color.}
 \vspace{-15pt}
\label{fig:MMFT}
\end{figure}
\vspace{-5pt}
\section{Introduction}
In the real world, acquiring knowledge requires processing multiple information sources such as visual, sound, and natural language individually and collectively. As humans, we can capture experience from each of these sources (like an isolated sound); however, we acquire the maximum knowledge when exposed to all sources concurrently. Thus, it is crucial for an ideal Artificial Intelligence (AI) system to process modalities individually and jointly.
One of the ways to understand and communicate with the world around us is by observing the environment and using language (dialogue) to interact with it \cite{lei2018tvqa}. 
A smart system, therefore, should be able to process visual information to extract meaningful knowledge as well as be able to use that knowledge to tell us what is happening. 
The story is incomplete if we isolate the visual domain from language. Now that advancements in both computer vision and natural language processing are substantial, solving problems demanding multimodal understanding (their fusion) is the next step. 
Answering questions about what can be seen and heard lies somewhere along this direction of investigation. In research towards the pursuit of combining language and vision, visual features are extracted using pre-trained neural networks for visual perception \cite{he2016deep, ren2015faster}, and word embeddings are obtained from pre-trained language models \cite{mikolov2013distributed, mikolov2013efficient, pennington2014glove, devlin2018bert} and these are merged to process multiple modalities for various tasks: visual question answering (VQA), visual reasoning, visual grounding. 

TVQA \cite{lei2018tvqa}, a video-based question answering dataset, is challenging as it provides more realistic multimodal question answers (QA) compared to other existing datasets . To answer TVQA questions, the system needs an understanding of both visual cues and language. In contrast, some datasets are focused either visually: MovieFIB \cite{maharaj2017dataset}, Video Context QA \cite{zhu2017uncovering}, TGIF-QA \cite{jang2017tgif}; or by language: MovieQA \cite{tapaswi2016movieqa}; or based on synthetic environments: MarioQA \cite{mun2017marioqa} and PororoQA \cite{kim2017deepstory}. 
 We choose TVQA because of its challenges.

The introduction of transformers \cite{vaswani2017attention} has advanced research in visual question answering and shows promise in the field of language and vision in general. 
Here, we adopt the pre-trained language-based transformer model, BERT \cite{devlin2018bert} to solve the VQA task. 
The human brain has vast capabilities and probably conducts processing concurrently. Like humans, an intelligent agent should also be able to process each input modality individually and collectively as needed. Our method starts with independent processing of modalities and the joint understanding happens at a later stage. Therefore, our method is one step forward toward better joint understanding of multiple modalities.
We use separate BERT encoders to process each of the input modalities namely Q-BERT, V-BERT and S-BERT to process question (Q), video (V),  and subtitles (S) respectively. 
Each BERT encoder takes an input source with question and candidate answer paired together. 
This is important because we want each encoder to answer the questions targeted at its individual source input. 
Thus, pairing up the question and candidate answers enables each stream to attend to the relevant knowledge pertinent to the question by using a multi-head attention mechanism between question words and a source modality. 
We then use a novel transformer based fusion mechanism to  jointly attend to aggregated knowledge from each input source, learning to obtain a joint encoding. 
In a sense, our approach is using two levels of question-to-input attention: first, inside each BERT encoder to select only relevant input; and second, at the fusion level, in order to fuse all sources to answer the common question. We show in our experiments that using Q-BERT, a separate BERT encoder for question and answer is helpful. 

\begin{figure*}
\begin{center}
 \includegraphics[width=\linewidth]{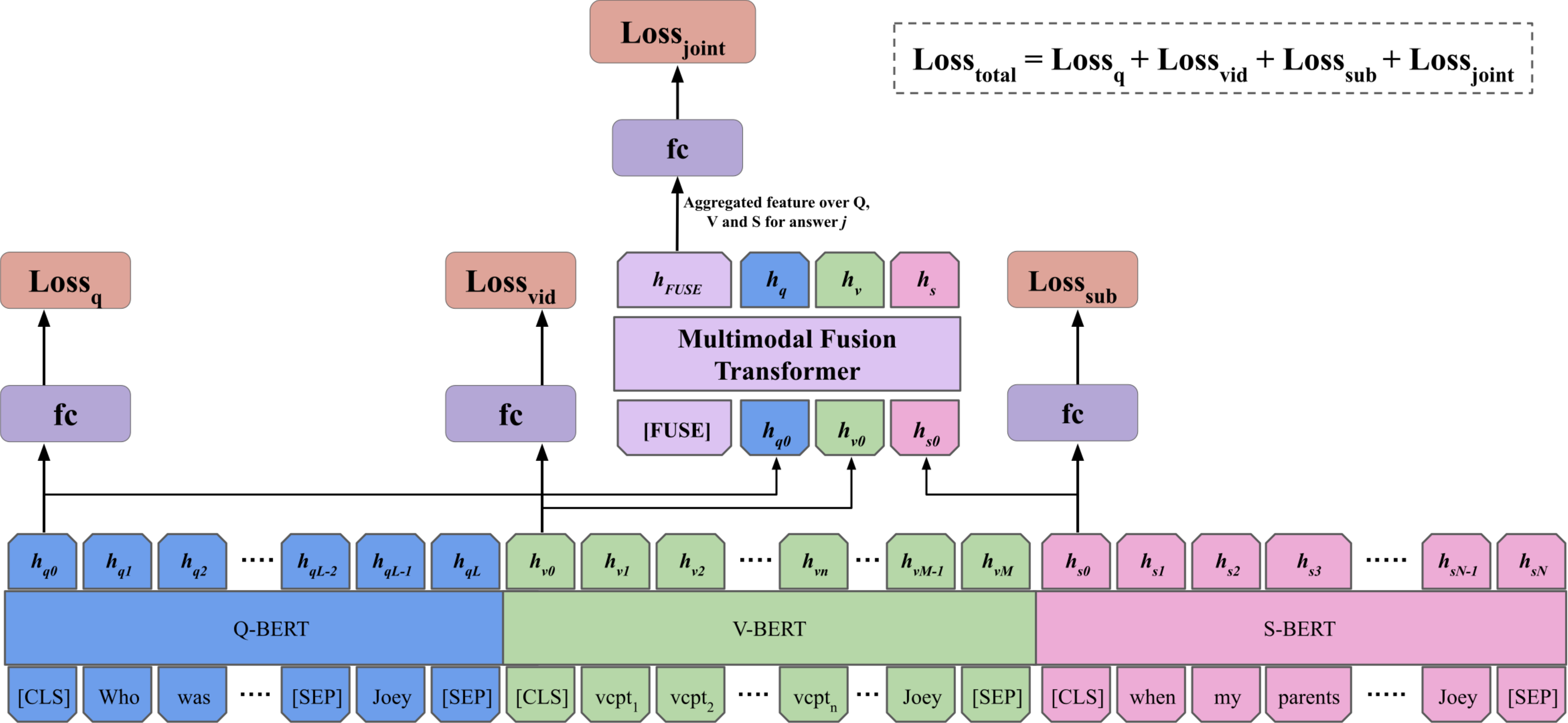}
 \vspace{-15pt}
\end{center}
  \caption{ Overview of the proposed approach. Q-BERT, V-BERT and S-BERT represent text encoder, visual encoder and subtitles encoder respectively. If $h_j$=Q+$A_j$ is $j_{th}$ hypothesis, then Q-BERT takes $h_j$, V-BERT takes visual concepts V$+h_j$, and S-BERT takes subtitles S$+h_j$ as inputs respectively. The aggregated features from each BERT are concatenated with [FUSE], a special trainable vector, to form a sequence and input into the MMFT module (see section \ref{fusion} for details). Outputs from the MMFT module for each answer choice are concatenated together and are input into a linear classifier to obtain answer probabilities. We optimize individual BERTs along with optimizing the full model together. $Loss_{total}$ denotes our objective function used to train the proposed architecture.  At inference time, we take features only from the MMFT module. 
  }
  \vspace{-15pt}
\label{fig:model}
\end{figure*}

\noindent Our contribution is three-fold: 

     \noindent \textbf{First}, we propose a novel multi-stream end-to-end trainable architecture which processes each input source separately followed by feature fusion over aggregated source features. Instead of combining input sources before input to BERT, we propose to process them individually and define an objective function to optimize multiple BERTs jointly. Our approach achieves state-of-the-art results on the video-based question answering task.
     
    \noindent \textbf{Second}, we propose a novel MultiModal Fusion Transformer (MMFT) module, repurposing transformers for fusion among multiple modalities. To the best of our knowledge, we are the first to use transformers for fusion.
    
    \noindent \textbf{Third}, we isolate a subset of visual questions, called TVQA-Visual (questions which require only visual information to answer them). Studying our method's behavior on this small subset illustrates the role each input stream is playing in improving the overall performance. We also present detailed analysis on this subset. 



%% file: Related.tex
\section{Related Work}
\noindent \textbf{Image-based Question Answering.}
Image-based VQA \cite{Yu_2015_ICCV, antol2015vqa, zhu2016visual7w, Jabri2016RevisitingVQ, Chao2018BeingNB} has shown great progress recently. A key ingredient is attention \cite{ilievski2016focused, chen2015abc, yu2017multi, yu2017multi-b, xu2016ask, anderson2018bottom}. 
Image based VQA can be divided based on the objectives such as generic VQA on real world images \cite{antol2015vqa, balanced_vqa_v2}, asking binary visual questions \cite{balanced_binary_vqa} and reasoning based VQA collecting visual information recurrently \cite{dmn, dmnplus, weston2014memory, memn2n, hudson2018compositional} to answer the question both in synthetic \cite{clevr, cog, nlvr} as well as real image datasets \cite{hudson2018gqa}.  

\noindent \textbf{Video-based Question Answering.}
Video-based QA is more challenging as it requires spatiotemporal reasoning to answer the question. 
\cite{lei2018tvqa} introduced a video-based QA dataset along with a two-stream model processing both video and subtitles to pick the correct answer among candidate answers. Some studies are: grounding of spatiotemporal features to answer questions \cite{lei2019tvqa}; a video fill in the blank version of VQA \cite{Mazaheri_2017_ICCV}; 
other examples include \cite{kim2019progressive, kim2019gaining, Zadeh_2019_CVPR, yi2019clevrer, mazaheri2018visual}.


\noindent \textbf{Representation Learning.}
BERT has demonstrated effective representation learning using self-supervised tasks such as masked language modeling and next sentence prediction tasks. The pre-trained model can then be finetuned for a variety of supervised tasks. QA is one such task. 
A single-stream approach takes visual input and text into a BERT-like transformer-based encoder; examples are: VisualBERT \cite{li2019visualbert}, VL-BERT \cite{su2019vlbert}, Unicoder-VL \cite{li2019unicoder} and B2T2 \cite{alberti2019fusion}.
Two-stream approaches need an additional fusion step;
ViLBERT \cite{lu2019vilbert} and LXMERT \cite{tan2019lxmert} employ two modality-specific streams for images. We take this a step further by employing three streams.
We use a separate BERT encoder for the question-answer pair. 
We are specifically targeting video QA and do not need any additional pre-training except using pre-trained BERT.

%% file: main_table.tex
\begin{table*}[t]
\footnotesize
\renewcommand{\arraystretch}{.9}
  \centering \setlength{\tabcolsep}{.6\tabcolsep}   
    \begin{tabular}{lllcccccccc}
       \toprule
& & & \multicolumn{8}{c}{\textbf{Input}}  \\
 \cmidrule{4-11} 
      & & & \multicolumn{2}{c}{Q} & \multicolumn{2}{c}{Q+V} & \multicolumn{2}{c}{Q+S} & \multicolumn{2}{c}{Q+V+S}  \\
       \cmidrule{4-11} 
   \multicolumn{1}{c}{\textbf{Method}} & \textbf{Text Feat} &
    \textbf{Vis. Feat} & w/o ts & w/ ts & w/o ts & w/ ts & w/o ts & w/ ts & w/o ts & w/ ts \\
       

        \midrule 
     LSTM(Q) & Glove & {-} & 42.74 & 42.74  & {-} & {-} & {-} & {-} & {-} & {-} \\
     MTL \cite{kim2019gaining} & Glove & cpt & {-} & {-}  & {-} & 43.45 & {-} & 64.36 & {-} & 66.22 \\
     Two-stream\cite{lei2018tvqa} & Glove & cpt & 43.50 & 43.50  & 43.03 & 45.03 & 62.99 & 65.15 & 65.46 & 67.70\\
     PAMN \cite{kim2019progressive} & Word2vec & cpt & {-} & {-} & {-} & {-} & {-} & {-} & 66.77 & {-}\\
     Single BERT & BERT & cpt & {-} & {-} & {-} & 48.95 & {-} & {-} & {-} & {72.20}\\
     STAGE \cite{lei2019tvqa} & BERT & reg & {-} & {-} & {-} & {-} & {-} & {-} & \textbf{68.56} & 70.50\\
       WACV20\cite{Yang_2020_WACV} & BERT & cpt & 46.88 & 46.88  & {-} & 48.95 & {-} & 70.65 & 63.07 & 72.45\\
    \midrule

    Ours-SF & BERT & cpt &\textbf{47.64} &  \textbf{47.64} & \textbf{49.52} & 50.65 & 69.92* & 70.33 & {65.55} & 73.10 \\
    Ours-MMFT & BERT & cpt & \textbf{47.64} & \textbf{47.64} & 49.32
    & \textbf{51.36} & \textbf{69.98}* & \textbf{70.79} & 66.10 &  \textbf{73.55}\\
                            Ours-MMFT(ensemble) & BERT & cpt & {-} & {-} & {-} & \textbf{53.08} & {-} & {-} & {-} &\textbf{74.97}\\
     
    \hline
    \end{tabular}
    \vspace{-5pt}
    \caption{\small Comparison of our method with baseline methods on TVQA validation set. STAGE uses regional features for detected objects in the video, all other models use visual concepts, ts= timestamp annotation, cpt=visual concepts, reg=regional features. Ours-SF represents proposed method with simple fusion, MMFT represents proposed multimodal fusion transformer, * indicates model trained with max\_seq\_len=512. The MMFT ensemble is 7x systems which use different training seeds.}
    \label{tab:main-results}
    \vspace{-12pt}
\end{table*} 

%% file: Approach.tex
\section{Approach}
Our approach  permits each stream 
to take care of the questions requiring only that input modality.
As an embodiment of this idea, we introduce the MultiModal Fusion Transformer with BERT encodings (MMFT-BERT) to solve VQA in videos. See fig. \ref{fig:MMFT} for the proposed MMFT module and fig. \ref{fig:model} for illustration of our full architecture.

\vspace{-5pt}
\subsection{Problem Formulation} \label{problem}
\vspace{-5pt}
In this work, we assume that each data sample is a tuple $(V, T, S, Q, A, l)$ comprised of the following:$V$: input video; $T$: $T=[t_{start}, t_{end}]$, i.e., start and end timestamps for answer localization in the video; $S$: subtitles for the input video; $Q$: question about the video and/or subtitles; $A$: set of C answer choices; $l$: label for the correct answer choice.

Given a question with both subtitles and video input, our goal is to pick the correct answer from C candidate answers. TVQA has 5 candidate answers for each question. Thus, it becomes a 5-way classification problem. 

\subsection{MultiModal Fusion Transformer with BERT encodings (MMFT-BERT)}
\subsubsection{Q-BERT:} \label{text-encoder}
Our text encoder named Q-BERT takes only QA pairs. The question is paired with each candidate answer $A_j$, where, $j={0, 1, 2, 3, 4};  |A|=C$. BERT uses a special token $[CLS]$ to obtain an aggregated feature for the input sequence, and uses $[SEP]$ to deal with separate sentences. We, therefore,  use the output corresponding to the $[CLS]$ token as the aggregated feature from Q-BERT and $[SEP]$ is used to treat the question and the answer choice as separate sentences.
The input I to the text encoder is formulated as:
\vspace{-5pt}
\begin{equation}
     I_{q_j} = [CLS] + Q + [SEP]  + A_j,
\end{equation}
\noindent where, $+$ is the concatenation operator, $[CLS]$ and $[SEP]$ are special tokens, $Q$ denotes the question, and  $A_j$ denotes the answer choice $j$, $I_{q_{j}}$ is the input sequence which goes into Q-BERT and represents the combination of question and the $j^{th}$ answer. We initiate an instance of the pre-trained BERT to encode each of the  $I_{q_j}$ sequences:
\vspace{-5pt}
\begin{equation}
    h_{{q_{0}}_j} = Q\text-BERT\hspace{1pt}(I_{q_{j}})\hspace{1pt}[\hspace{1pt}0\hspace{1pt}],
\end{equation}
\noindent where [0] denotes the index position of the aggregated sequence representation for only textual input. Note that, the [0] position of the input sequence is $[CLS]$. 

\subsubsection{V-BERT:} \label{visual-encoder}
We concatenate each QA pair with the video to input to our visual encoder V-BERT. V-BERT is responsible for taking care of the visual questions. Pairing question and candidate answer with visual concepts allows V-BERT to extract visual knowledge relevant to the question and paired answer choice.
Input to our visual encoder is thus formulated as follows:
\vspace{-5pt}
\begin{equation}
     I_{v_j} = [CLS] + V \hspace{3pt} + "." + \hspace{3pt} Q \hspace{3pt}+ [SEP] + \hspace{3pt} A_j,
\end{equation}

\noindent where, 
$V$ is the sequence of visual concepts\footnote{Visual concepts is a list of detected object labels using FasterRCNN \cite{ren2015faster} pre-trained on Visual Genome dataset. We use visual concepts provided by \cite{lei2018tvqa}.}, "." is used as a special input character, $I_{v_j}$ is the input sequence which goes into our visual encoder.
\vspace{-5pt}
\begin{equation}
    h_{{v_0}_j} = V\text-BERT\hspace{1pt}(I_{v_j})\hspace{1pt}[\hspace{1pt}0\hspace{1pt}],
\end{equation}

\noindent where, [0] denotes the index position of the aggregated sequence representation for visual input.

\subsubsection{S-BERT:} \label{sub-encoder}
The S-BERT encoder applies attention 
between each QA pair and subtitles and results in an aggregated representation of subtitles and question for each answer choice.
Similar to the visual encoder, we concatenate the QA pair with subtitles as well; and the input is:
\vspace{-5pt}
\begin{equation}
     I_{s_j} = [CLS] + S \hspace{3pt}+ "." + \hspace{3pt} Q \hspace{3pt} + [SEP] \hspace{3pt}+ A_j,
\end{equation}

\noindent where, 
$S$ is the subtitles input, 
$I_{s_j}$ is the resulting input sequence which goes into the S-BERT encoder.
\vspace{-5pt}
\begin{equation}
    h_{{s_0}_j} = S\text-BERT\hspace{1pt}(I_{s_j})\hspace{1pt}[\hspace{1pt}0\hspace{1pt}].
\end{equation}

\noindent where, [0] denotes the index position of the aggregated sequence representation for subtitles input.

\subsubsection{Fusion Methods} \label{fusion}
Let ${I}_i \in R^d$ denote the feature vector for $i_{th}$ input modality with total $n$ input modalities i.e.\ $I_1, I_2, ..., I_n$, $d$ represents the input dimensionality. We discuss two possible fusion methods:

\noindent \textbf{Simple Fusion:}
 A simple fusion method is a Hadamard product between all input modalities and given as follows:
 \vspace{-5pt}
\begin{equation}
    h_{FUSE} = {I}_1 \odot {I}_2 \odot ... \odot {I}_n,
\end{equation}
where, $h_{FUSE}$ is the resulting multimodal representation which goes into the classifier.
Despite being extremely simple, this method is very effective in fusing multiple input modalities.

\vspace{5pt}
\noindent \textbf{MultiModal Fusion Tranformer (MMFT):}
The MMFT module is illustrated in fig. \ref{fig:MMFT}.
We treat ${I}_i$ as a fixed $d$-dimensional feature aggregated over input for modality $i$. 
Inspired by BERT\cite{devlin2018bert}, we treat aggregated input features from multiple modalities as a sequence of features by concatenating them together. We concatenate a special trainable vector $[FUSE]$\footnote{ $[FUSE]$ is initialized as a $d$-dimensional zero vector.} 
as the first feature vector of this sequence. The final hidden state output corresponding to this feature vector is used as the aggregated sequence representation over input from multiple modalities denoted as $h_{FUSE}$. 
\begin{equation}
     h_{FUSE} = MMFT({I}_1 + {I}_2 + ... + {I}_n)\hspace{1pt}[\hspace{1pt}0\hspace{1pt}],
\end{equation}

\noindent where, $+$ 
is the concatenation operator,
[0] indicates the index position of the aggregated sequence representation over all input modalities.

In our case, we have three input types: QA pair, visual concepts and subtitles. For inputs $i=\{1,2,3\}$ 
and answer index $j=\{0,1,2,3,4\}$, the input to our MMFT module is  ${I}_1 = h_{{q_0}_j}$,  ${I}_2 = h_{{v_0}_j}$,  and  ${I}_3 = h_{{s_0}_j}$ and the output is $h_{FUSE}$ denoting hidden output corresponding to the [FUSE] vector. 
Here,  $h_{{q_0}_j}$, $h_{{v_0}_j}$, and $h_{{s_0}_j}$ are the aggregated outputs we obtain from Q-BERT, V-BERT and S-BERT respectively.

\subsubsection{Joint Classifier} \label{joint-classifier}
Assuming a hypothesis for each tuple $(V, T, S, Q, A_j)$, where $A_j \in A ; j=0,..,4$ denotes five answer choices, our proposed Transformer Fusion module outputs $h_{{FUSE}_j} \in R^d$.
We concatenate the aggregated feature representation for all the answers 
together and send this to  a joint classifier to produce 5 answer scores, as follows:
\vspace{-5pt}
\begin{equation}
    h_{final} = h_{FUSE_0} +
                h_{FUSE_1} +
               ... +
               h_{FUSE_4},
               \vspace{-5pt}
\end{equation}
\vspace{-10pt}
\begin{equation}
\vspace{-5pt}
    scores_{joint} = classifier_{joint}(h_{final}),
\end{equation}

\noindent where, $h_{final} \in R^{C \cdot d}$ and $scores_{joint} \in R^C$, C denotes number of classes. 

\vspace{-5pt}
\subsection{Objective Function} \label{objective}
Along with joint optimization, each of the Q-BERT, V-BERT and S-BERT are optimized with a single layer classifier using a dedicated loss function for each of them.
Our objective function is thus composed of four loss terms: one each to optimize each of the input encoders Q-BERT, V-BERT and S-BERT, and a joint loss term over classification using the combined feature vector. 
The formulation of the final objective function is as follows:
\begin{equation}
    L_{total} = L_{q} + L_{vid} + L_{sub} + L_{joint},
\end{equation}

\noindent where, $L_{q}$, $L_{vid}$, $L_{sub}$, and $L_{joint}$ denote loss functions for question-only, video, subtitles, and joint loss respectively;
all loss terms are computed using softmax cross-entropy loss function using label $l$. The model is trained end-to-end using $L_{total}$.

%% file: test-public.tex

\begin{table}
\footnotesize

\centering \setlength{\tabcolsep}{.56\tabcolsep} 
\begin{tabular}{lll}
\toprule
\textbf{Input} & \textbf{Model} & \textbf{Acc (\%)}\\

\hline
\multirow{3}{*}{Q+V}  & MTL \cite{kim2019gaining} & 44.42 \\
                    & Two-stream \cite{lei2018tvqa} &  45.44\\
                     
                     & Ours - MMFT & \textbf{51.83}\\
 \hline
    \multirow{5}{*}{Q+V+S}  & MTL \cite{kim2019gaining}  & 67.05 \\
                    & Two-stream \cite{lei2018tvqa} & 68.48 \\
                     & STAGE \cite{lei2019tvqa}  & 70.23 \\ 
                     & WACV20 \cite{Yang_2020_WACV}  & 72.71 \\
                     & Ours - MMFT model  & \textbf{72.89}\\
\hline
\end{tabular}

\caption{\small Performance comparison of different models on TVQA testset-public with timestamp annotations. All models use visual concepts except STAGE. We do not report numbers for other comparisons (Q+S and w/o ts) because only limited attempts are allowed to the test server for evaluation.}\label{tab:test-public}
\end{table}

\begin{table}[t]
\scriptsize
\renewcommand{\arraystretch}{.9}
  \centering 
  \setlength{\tabcolsep}{.35\tabcolsep}   
    \begin{tabular}{llccccccc}
       \toprule

        Inp. &  \multicolumn{1}{c}{Method}  & \multicolumn{7}{c}{Question family (Accuracy\%)}  \\
        \cmidrule{3-9} 
       
    &   & what & who & where & why & how & others & all \\
        \midrule 

 \multirow{4}{*}{Q+V}  &  Two-stream  & 47.70 & 34.60 & 47.86 & 45.92 & 42.44 & 39.10 & 45.03 \\
 &  WACV20 & 51.31 & 41.14 & 52.86 & 48.45 & 46.24 & 36.86 & 48.95 \\

     &  Ours-SF & 52.76 & 42.52 & 52.36 & 51.42 & 46.75 & 41.61 & 50.65\\
    &   Ours-MMFT & \textbf{52.97} & \textbf{43.58} & \textbf{54.00} & \textbf{53.00} & \textbf{46.97} & \textbf{44.16} & \textbf{51.36}\\
    \midrule
  \multirow{3}{*}{Q+V+S}   &  Two-stream  & 66.05 & 67.99 & 61.46 & 71.53 & 78.77 & 74.09 & 67.70\\
    &  Ours-SF & 71.52 & 72.10 & 68.93 & \textbf{76.99} & \textbf{82.25} & \textbf{83.2} & 73.10\\
   &  Ours-MMFT & \textbf{72.22} & \textbf{72.39} & \textbf{69.89} & 76.92 & 81.74 & 82.48 & \textbf{73.55}\\
     
    \hline
    \end{tabular}
    \caption{\small Performance comparison for each question family. All models are trained with localized input (w/ ts).}
    \label{tab:q-wise-stats}
    \vspace{-1pt}
\end{table} 
\begin{figure}[t]
\begin{center}
 \includegraphics[width=\linewidth]{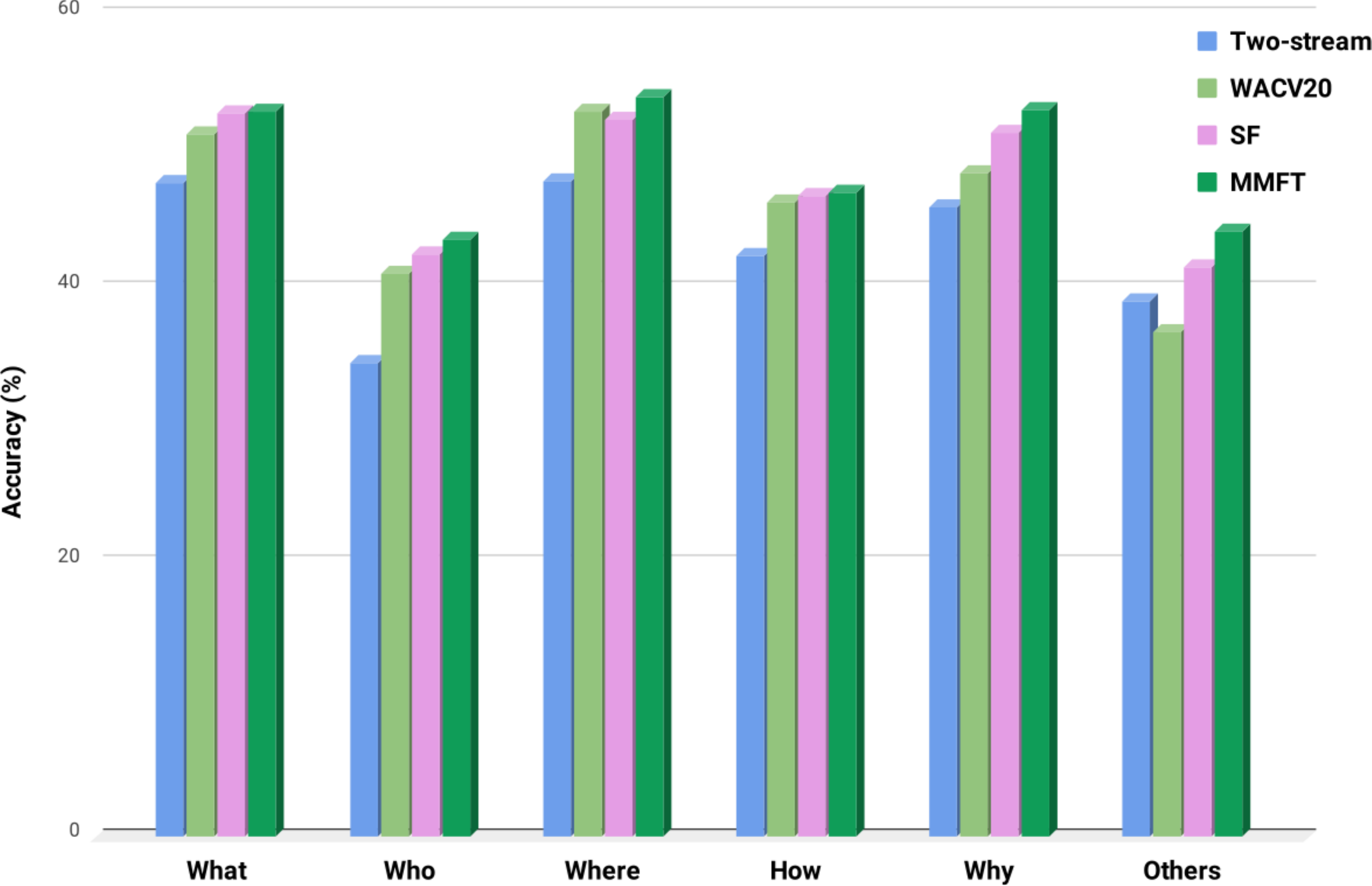}
  \vspace*{-17pt}
\end{center}
  \caption{\small Accuracy comparison with respect to the question family between Two-stream \cite{lei2018tvqa}, WACV20 \cite{Yang_2020_WACV} and our method on the validation set of TVQA. Models were trained on Q+V. MMFT outperforms on all question types for Q+V.}
\label{fig:question-families}
\vspace{-5pt}
\end{figure}
\begin{figure}[t]
\centering
 \includegraphics[width=\linewidth]{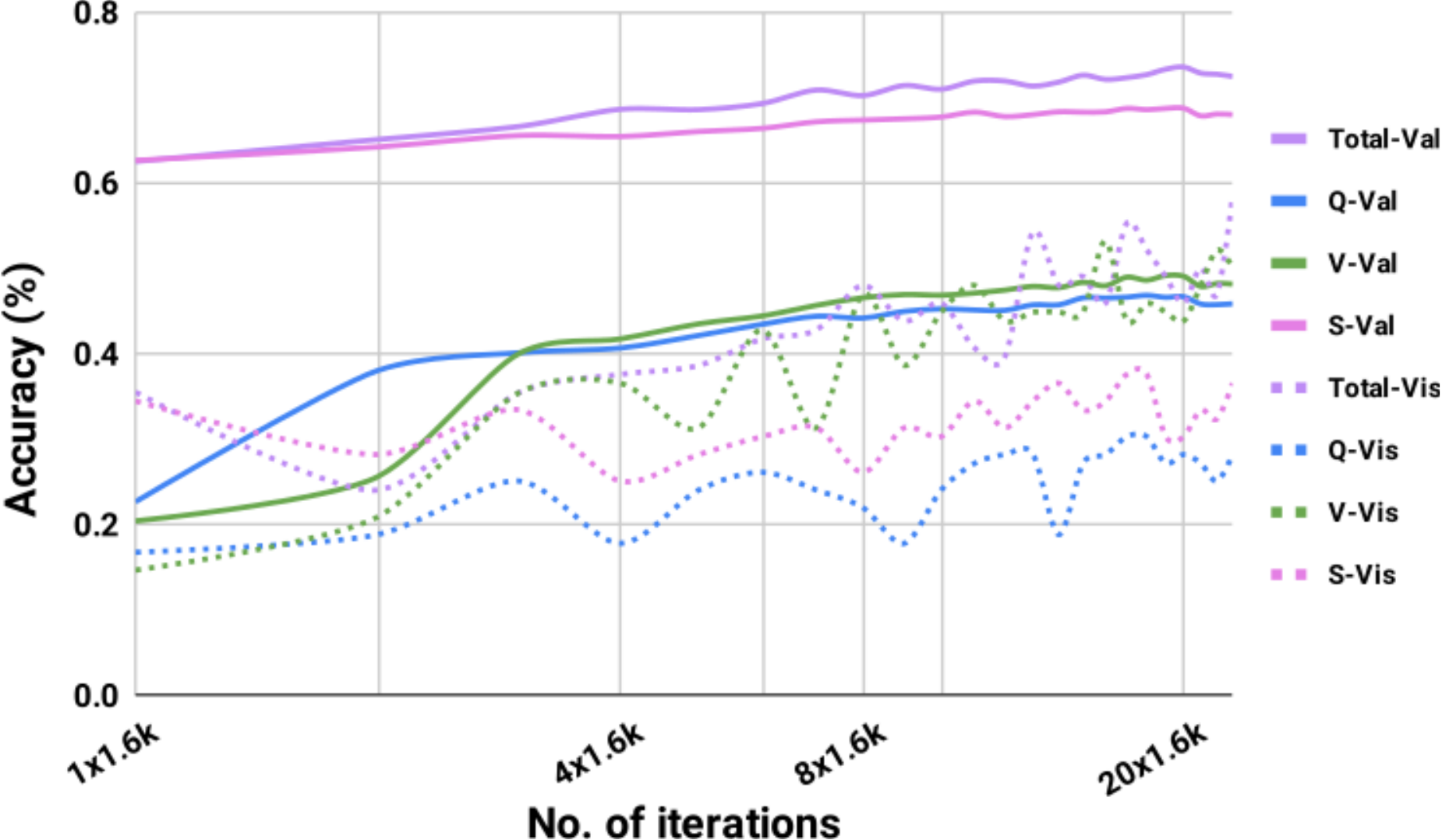}
 \vspace*{-15pt}

  \caption{\small Testing accuracy curves on TVQA-Vis. (clean) and TVQA val set for different BERT streams during training. Solid lines: validation accuracy, dotted lines: visual set accuracy. Although S-BERT is significantly above V-BERT for full validation set, for visual set, we can see that V-BERT is well above Q-BERT and S-BERT. This shows that each BERT contributes to the questions it is responsible for. Numbers are log-scaled.}
\label{fig:visual-accuracy}
\vspace{-18pt}
\end{figure}

\begin{table}[t]
\footnotesize
\renewcommand{\arraystretch}{.9}
  \centering \setlength{\tabcolsep}{.56\tabcolsep}   
    \begin{tabular}{lll}
       \toprule

           \textbf{Method}  & \textbf{TVQA-Vis. (clean)} & \textbf{TVQA Vis. (full)} \\

    \midrule
    
      Two-stream & 35.42 & 29.49\\
      WACV20 & 42.71 & \textbf{40.00}\\
      SF & 42.71 & 34.37 \\
      MMFT & \textbf{46.88} & 39.57 \\
    \hline
    \end{tabular}
    \vspace{-5pt}
    \caption{\small Performance comparison on TVQA-Visual questions for clean set and full set. Numbers reported for only Q+V (w/ts) model. Numbers are reported as percentage.}
    \label{tab:visual-set-performance}
    \vspace{-15pt}
\end{table} 

%% file: TablesAndFigures.tex
\begin{figure*}[t]
\begin{center}
 \includegraphics[width=\linewidth]{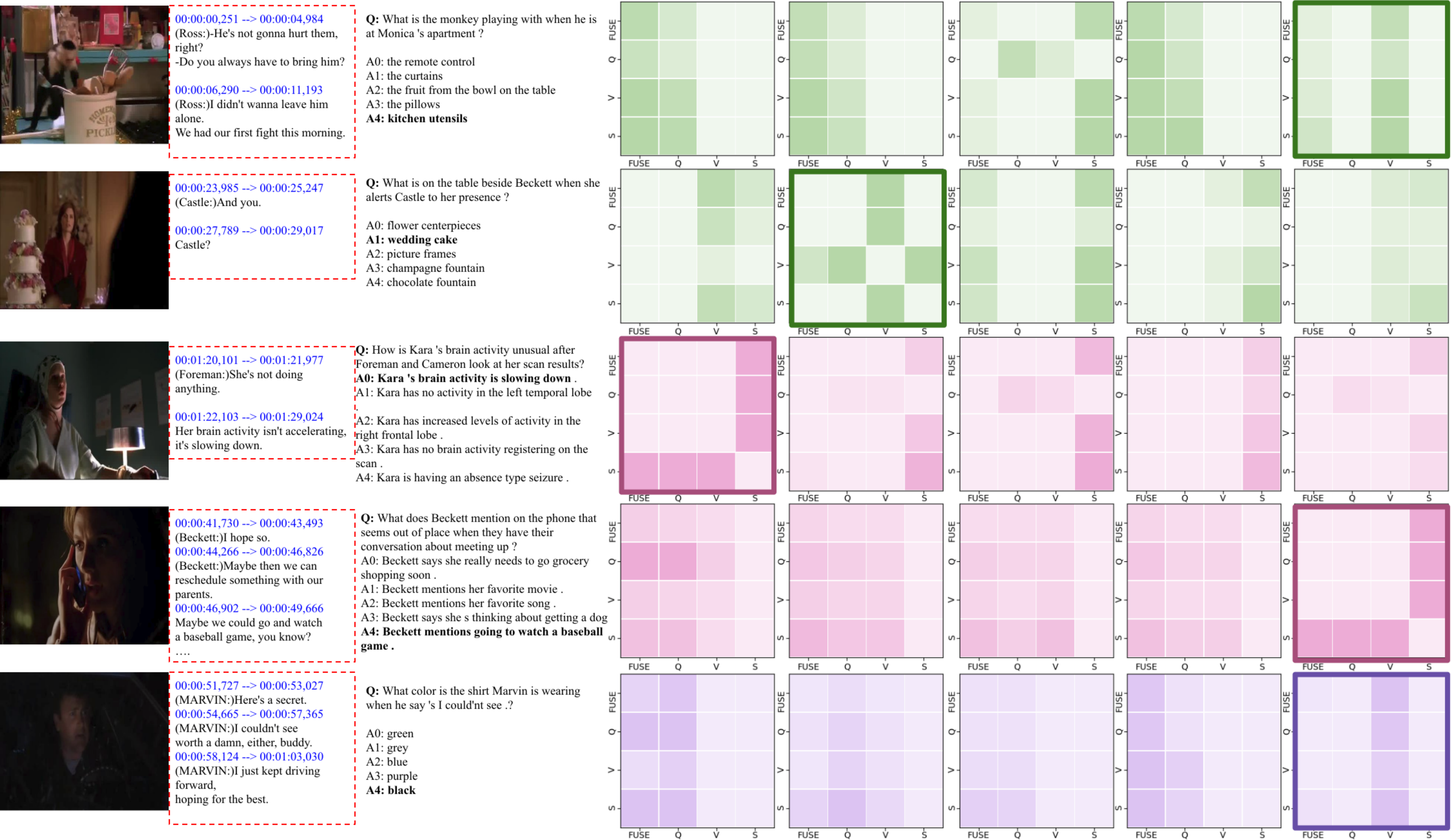}
 
\end{center}
  \caption{Visualization of multi-head attention (averaged over all heads) between different source features: Q, V and S for our best model. MMFT takes a sequence: [FUSE, Q, V, S] 
  and uses multi-head attention for multimodal fusion. [FUSE] is the aggregated feature over Q, V, and S; column 1: representative image frame, column 2: localized subtitles, column 3: question with candidate answers (correct answer with corresponding attention map is in bold text and box respectively), columns 4-8 show attention for A0, A1, A2, A3, and A4 respectively. Top 2 rows show attention weights for visual questions, next 2 rows are subtitles-based questions. Last row, depends on both subtitles and visual information. See sec. \ref{head-attn} and supplementary work for details and insights.}
\label{fig:attention-maps}
\vspace{-15pt}
\end{figure*}

%% file: Experimental.tex
\vspace{-5pt}
\section{Dataset}

In TVQA, each question (Q) has 5 answer choices. 
It consists of ~152K QA pairs with 21.8K video clips. 
Each question-answer pair has been provided with the localized video V to answer the question Q, i.e., start and end timestamps are annotated. Subtitles S have also been provided for each video clip. 
See supplementary work for a few examples.
\vspace{-5pt}
\subsection{TVQA-Visual}
\vspace{-5pt}
To study the behavior of state-of-the-art models on questions where only visual information is required to answer the question correctly,
we selected 236 such visual questions. Due to imperfections in the object detection labels, only approximately 41\% of these questions have the adequate visual input available. 
We, therefore, refer to TVQA-Visual in two settings: \textbf{TVQA-Visual (full):}-- full set of 236 questions. A human annotator looked into the video carefully to ensure that the raw video is sufficient to answer the question without using subtitles.
\textbf{TVQA-Visual (clean):} This is the subset of 96 questions where the relevant input was available, yet the models perform poorly. For this subset, we rely on a human annotator's judgement who verified that either the directly related visual concept or the concepts hinting toward the correct answer are present in the list of detected visual concepts. 
For instance, if the correct answer is \say{kitchen}, either \say{kitchen} or related concepts (e.g. \say{stove}, \say{plate}, \say{glass}, etcetera) should be present in the list of visual concepts. 
Thus, this easier subset is termed as TVQA-Visual (clean). TVQA-visual, although small, is a diagnostic video dataset for systematic evaluation of computational models on spatio-temporal question answering tasks and will help in looking for ways to make the V-stream contribution more effective.
See supplementary material for the distribution of visual questions based on reasons for failure.
If a model is correctly answering TVQA-visual questions which are not \say{clean} (the relevant concepts are missing from the visual input), that is because of statistical bias in the data.

\section{Experiments and Results}

\subsection{Baselines}
\noindent \textbf{LSTM(Q):} 
LSTM(Q) is a BiLSTM model to encode question and answer choices.
The output from LSTM for question and each answer choice is concatenated and is input to a 5-way classifier to output 5 answer probability scores. 

\noindent \textbf{MTL:} MTL \cite{kim2019gaining} uses two auxiliary tasks with the VQA task: temporal alignment and modality alignment. 

\noindent \textbf{Two-stream:} \cite{lei2018tvqa} uses two separate streams for attention-based context matching each input modality with question and candidate answers. A BiLSTM with max pooling over time is used to aggregate the resulting sequence. 

\noindent \textbf{BERT:} A single pre-trained BERT instance is finetuned on QA pair along with visual concepts and subtitles all together (Q+V+S). 

\noindent \textbf{STAGE:} \cite{lei2019tvqa} uses moment localization and object grounding along with QA pair and subtitles. STAGE uses BERT features to encode text and spatio-temporal features for video.

\noindent \textbf{WACV20:} \cite{Yang_2020_WACV} concatenates subtitles and visual concepts with QA pairs and input to BERT along with late fusion for Q+V and Q+S.

\vspace{-5pt}
\subsection{MMFT-BERT}
For video representation, we use detected attribute object pairs as visual features provided by \cite{lei2018tvqa}. We follow \cite{lei2018tvqa} and only unique attribute-object pairs are kept. Q-BERT, V-BERT and S-BERT are initialized with $BERT_{base}$ pre-trained on lower-cased English text with masked language modeling task. The MMFT module uses single transformer encoder layer (L=1) with multi-head attention. We use 12 heads (H=12) for multi-head attention in the MMFT module for our best model. We initialize the MMFT module with random weights. A \textit{d}-dimensional hidden feature output corresponding to [CLS] token is used as an aggregated source feature from each BERT. We concatenate these aggregated features for each candidate answer together to acquire a feature of size $5\cdot d$. A 5-way classifier is then used to optimize each of Q-BERT, V-BERT and S-BERT independently. 
For joint optimization of the full model, we treat the encoders' output as a sequence of features with the order $[[FUSE], h_{{q_0}_j}, h_{{v_0}_j}, h_{{s_0}_j}]$ and input this into the MMFT module (\textit{[FUSE]} is a trainable \textit{d}-dimensional vector parameter). Output corresponding to \textit{[FUSE]} token is treated as an accumulated representation $h_{{FUSE}_j}$ over all input modalities for answer \textit{j}. We concatenate  $h_{{FUSE}_j}$ for each answer choice to obtain $h_{final}$ for the joint classification.
We learn four linear layers, one on top of each of the three input encoders and the MMFT encoder respectively. Thus, each linear layer takes a ($5 \cdot d$)-dimensional input and produces 5 prediction scores. 

\noindent \textbf{Training Details.} The entire architecture was implemented using Pytorch \cite{NIPS2019_9015} framework. All the reported results were obtained using the Adam optimizer \cite{kingma2014adam} with a minibatch size of 8 and a learning rate of 2e-5. Weight decay is set to 1e-5. All the experiments were performed under CUDA acceleration with two NVIDIA Turing (24GB of memory) GPUs. In all experiments, the recommended train / validation / test
split was strictly observed. We use the 4th last layer from each BERT encoder for aggregated source feature extraction. 
The training time varies based on the input configuration. It takes $\sim$4 hrs to train our model with Q+V and $\sim$8-9 hrs to train on the full model for a single epoch. All models were trained for 10 epochs. Our method achieves its best accuracy often within 5 epochs.
\vspace{-5pt}
\subsection{Results}
All results here use the following hyperparameters: input sequence length max\_seq\_len=256, \# heads H=12, \# encoder layers L=1 for the MMFT module, and pre-trained $BERT_{base}$ weights for Q-BERT, V-BERT and S-BERT unless specified explicitly.

\noindent \textbf{With timestamp annotations (w/ ts).}
Columns with \say{w/ ts} in table \ref{tab:main-results} show results for input with timestamp localization. We get consistently better results when using localized visual concepts and subtitles. We get ~1.7\%  and 0.65\% improvement over WACV20 \cite{Yang_2020_WACV} with simple fusion for Q+V and Q+V+S inputs respectively. When using the MMFT for fusion, our method achieves SOTA performance with all three input settings: Q+V ($\uparrow$ 2.41\%), Q+S ($\uparrow$ 0.14) and Q+V+S ($\uparrow$ 1.1\%) (see table  \ref{tab:main-results}). Our fusion approach contributes to improved performance and gives best results for localized input. See table \ref{tab:q-wise-stats} and fig. \ref{fig:question-families} for results w.r.t. question family.

\noindent \textbf{Without timestamp annotations (w/o ts). }
We also train our model on full length visual features and subtitles. Our method with simple fusion and MMFT on Q+V input outperforms Two-stream \cite{lei2018tvqa} by absolute 6.49\% and 5.59\% with simple fusion and MMFT respectively. 
We truncate the input sequence if it exceeds max\_seq\_len. Subtitles without timestamps are very long sequences (~49\% of subtitles are longer than length 256), hence QA pair might be truncated. Thus, we rearrange our input without timestamps as follows: \say{$Q \hspace{2pt} [SEP] \hspace{2pt} A_j \hspace{3pt}. \hspace{3pt} V$} and \say{$Q \hspace{2pt} [SEP] \hspace{2pt} A_j \hspace{3pt}. \hspace{3pt} S$} for V-BERT and S-BERT respectively. Models with Q+S input are trained with max\_seq\_len=512 and Q+V+S models are trained with max\_seq\_len=256 due to GPU memory constraints. For Q+S and Q+V+S, we observe 69.92\%  and 65.55\% with simple fusion, using MMFT produces 69.98\% and 66.10\% val. accuracy respectively. 

\noindent \textbf{Results on test set.}
 \label{test-public} 
 TVQA test-public set does not provide answer labels and requires submission of the model's predictions to the evaluation server. Only limited attempts are permitted. The server's evaluation results are shown in table \ref{tab:test-public}.  
 MMFT improves results by ($\uparrow 6.39\%$) on Q+V. For Q+V+S, WACV20 reported 73.57\% accuracy with a different input arrangement than MMFT. When compared with the model with the same input, MMFT performs slightly better ($\uparrow 0.17\%$). Due to limited chances for submission to the test server for evaluation, the reported accuracy for Q+V+S is from one of our earlier models, not from our best model. 
\vspace{-5pt}
\begin{table}
\footnotesize
\centering
\begin{tabular}{lll}
\toprule
& \textbf{Model} & \textbf{Acc.(\%)} \\ \hline
1 & Single BERT  & 72.20  \\
2 & Ours Simple Fusion, Single Loss & 71.82 \\
3 & Ours Simple Fusion, FO & 73.10  \\
4 & MMFT w/ BERT Encoder freezed & 57.94 \\
5 & Ours Simple Fusion, FO, +Img & 71.82 \\
6 & MMFT-BERT L=2, H=12 & 72.61  \\
7 & MMFT-BERT L=2, H=12 w/ skip &  72.62\\
8 & MMFT-BERT L=1, H=1 & 72.66\\
9 & MMFT-BERT L=1, H=12 & \textbf{73.55}\\

\hline
\end{tabular} 
\caption{ Ablations over the design choices for the proposed architecture. L = no. of encoder layers in MMFT module, H = no. of heads in MMFT module, +Img = Resnet101 features pooled over video frames.
Rows 3-9 are trained with our full objective (FO). All models are trained for Q+V+S with timestamp annotations. 
}\label{tab:ablation}
\vspace{-15pt}
\end{table}

\subsubsection{Model Analysis}
\noindent \textbf{Performance analysis on TVQA-Visual.}
To study the models, we evaluate Two-stream \cite{lei2018tvqa}, WACV20 \cite{Yang_2020_WACV} and our method on both TVQA-Visual (full) and TVQA-Visual (clean). See table \ref{fig:visual-accuracy} for full results.

\noindent \textbf{TVQA-Visual (full):} Our method outperforms Two-stream \cite{lei2018tvqa} by 10.08\% but drops by 0.43\% compared to WACV20 \cite{Yang_2020_WACV}. TVQA-Visual (full) has approximately 59\% of the questions with missing visual concept or require extra visual knowledge. All three models including ours were trained on visual concepts. Inadequate input, therefore, makes it difficult for the models to attend the missing information.

\noindent \textbf{TVQA-Visual (clean):}
We observe $(\uparrow11.46\%)$ and  $(\uparrow4.17\%)$ improvement for clean set compared to Two-stream and WACV20.  
TVQA-Visual (clean) has relevant visual concepts or related concepts to the answer present in the input. Yet, it is challenging for existing methods (including ours) to perform well. Although our model observes significant improvement ($\uparrow$4-11\%) over baselines for this experiment, the take away message is that it is not enough. This subset of TVQA, therefore, serves as a good diagnostic benchmark to study the progress of exploiting visual features for multimodal QA tasks. Fig. \ref{fig:visual-accuracy} visualizes the test performance of each stream on TVQA-Visual clean during training our Q+V model on TVQA. 

\noindent \textbf{Performance analysis w.r.t multimodal attention.}
 \label{head-attn}
We study the behavior of the MMFT module for aggregating multimodal source inputs (Q, V, and S), we take our best model trained on all three sources, and evaluate it on questions which need knowledge about either the visual world, dialogue or both. We then visualize the average attention score map over all heads inside MMFT module (H=12) for each candidate answer, see fig. \ref{fig:attention-maps}. Top 2 rows show attention scores computed among all 3 input sources and the [FUSE] vector for visual questions. Since, [FUSE] is the aggregated output over all input modalities. For instance, visual part should contribute more if the question is about the visual world. We can see the attention map for the correct answer has high attention scores between V and [FUSE] vector. The incorrect answers attend to the wrong sources (either Q or S). Similar is the behavior for rows 3-5, where the question is about subtitles, and the correct answer gives most weight to the subtitles compared to the incorrect answers. Heatmaps for incorrect answers are either focused more on a wrong single input source or the combination of them.

\noindent \textbf{Positional Encodings for V-BERT.}
\label{pos-encoding-vbert}
Positional encoding is done internally in BERT. When finetuned, for V-BERT, the positional encoding has no effect. This has been verified by training our Q+V model with simple fusion (Ours-SF), where the input to V-BERT is a shuffled sequence of objects; no drastic difference was observed (shuffled: 50.32\% vs. not shuffled: 50.65\%).

\subsubsection{Ablations}
All ablations were done with Q+V+S input. See table \ref{tab:ablation} for complete results.

\noindent \textbf{Simple fusion vs. MMFT}
Though using simple fusion for combining multimodal inputs is very effective and already outperforms all of the baselines, it lacks the basic functionality of explainability. Using MMFT instead, not only gives us an improvement ($\uparrow$0.71\% for Q+V and $\uparrow$0.39\% for Q+V+S) over simple fusion, but is also more explainable. 

\noindent \textbf{Single loss vs. multiple losses.}
A simple design choice could be to use just joint loss instead of multiple loss terms. However, through our experiments, we find that using single joint loss term hurts the performance (71.82\%). Optimizing each BERT along with optimizing the full model jointly gives us best results (73.10\%) even without using MMFT. 

\noindent \textbf{Single head vs. multi-head MMFT.}
In an attempt to know if simplicity (single head) has an advantage over using multi-head attention, we trained MMFT-BERT with H=1. Using single head attention for fusion consistently performed lower than using multiple heads (72.66\% vs. 73.55\%) (we set H=12). Our hypothesis is that since pre-trained BERTs have 12 heads, attention within each source BERT was local ($d_{model}$/H). Using single head attention over features which were attended in a multi-head fashion may be hurting the features coming out of each modality encoder. Thus, it makes more sense to keep the attention local inside MMFT if the input encoders use local attention to attend to input sequences. 

\noindent \textbf{Single layer fusion vs. stacked fusion.} 
Another design parameter is \#encoder layers (L) in MMFT. We trained our full model with three settings: a) single encoder layer L=1, b) stacked encoder layer L=2, and c) stacked encoder with skip connection. a) gives best results (73.55\%), whereas both b) and c) fusion hurts (72.61\% and 72.62\%). 
Note that all variants of our models are slightly better in performance than our baseline methods.  

\noindent \textbf{Resnet features vs. visual concepts.}
To study if incorporating additional visual context is advantageous, we experimented with Resnet101 features for visual information. We used Resnet101 features pooled over time along with visual concepts. We used question-words-to-region attention for aggregating visual features; adding this aggregated visual feature to Ours-SF hurts the performance (71.82\%); using object labels was consistently more useful than visual features in various other experimental settings.


%% file: Conclusion.tex
\section{Conclusion}
Our method for VQA uses multiple BERT encodings to process each input type separately with a novel fusion mechanism to merge them together. We repurpose transformers for using attention between different input sources and aggregating the information relevant to the question being asked. Our method outperforms state-of-the-art methods by an absolute $\sim$2.41\% on Q+V and $\sim$1.1\% on Q+V+S on TVQA validation set.
Our proposed fusion lays the groundwork to rethink transformers for fusion of multimodal data in the feature dimension. 

%% file: supplementary.tex


\subsection{Improved comprehension of submitted paper}
Certain parts of the submitted paper will be more clear to the reader if s/he is familiar with the concepts explained in \cite{vaswani2017attention} and \cite{devlin2018bert}. For instance, the attention mechanism illustrated in the submitted paper's figure 1 needs understanding of transformers \cite{vaswani2017attention}.
\subsection{TVQA-Visual}
See figure \ref{fig:visual-set-stats} for some statistics about TVQA-Visual set. Almost 59\% of the questions have not enough input available (plot shows results for 200 questions, rest of the 35 questions are "who" questions and need character recognition). We will make this list of questions available to the community for further research.

\subsection{Experiments and Results} \label{results}
\noindent \textbf{Evaluation Metric.} Multiple choice question answering accuracy is used to evaluate each model in this work.

 \noindent \textbf{Further discussion about results without timestamp annotations.}
 For experiments without timestamp annotations, for Q+V and Q+S, the only competitor whose results are available is Two-stream \cite{lei2018tvqa}; in these categories, MMFT is more than 6-7\% better than the Two-stream, where, for Q+S, we train MMFT with a sequence length of 512. For Q+V+S, MMFT achieves 66.10\% with max\_seq\_len=256. STAGE reports 2.46\% higher accuracy. We had a GPU memory limitation and could only train our model with input size of 256. Had we had access to at least 4 GPUs (24GB of memory), we would have been able to train our full model with input size of 512, which would have presumably given us a similar boost we witnessed for Q+S without timestamps (Q+S is $\sim$3\% better than Q+V+S). Therefore, we believe our model would perform better when provided with increased input length.
\label{wo_ts}
\begin{figure}[t]
\centering
\hspace{-17pt} \includegraphics[width=\linewidth]{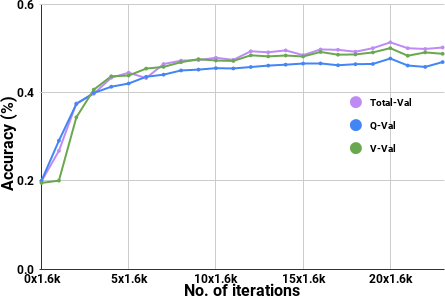}
\vspace*{-5pt}
 \caption{Validation accuracy curves for Q-BERT, V-BERT and full model when trained on Q+V input. Although Q-BERT performs lower than V-BERT as expected, it helps when Q-BERT is kept as a separate stream. During initial training, Q-BERT trains quickly than V-BERT. After first epoch, V-BERT starts outperforming Q-BERT as the model learns to leverage visual stream to answer the questions.}
\label{fig:qbert-vs-vbert}
\vspace{-10pt}
\end{figure}

\begin{figure*}[t]
\centering
 \includegraphics[width=0.28\linewidth]{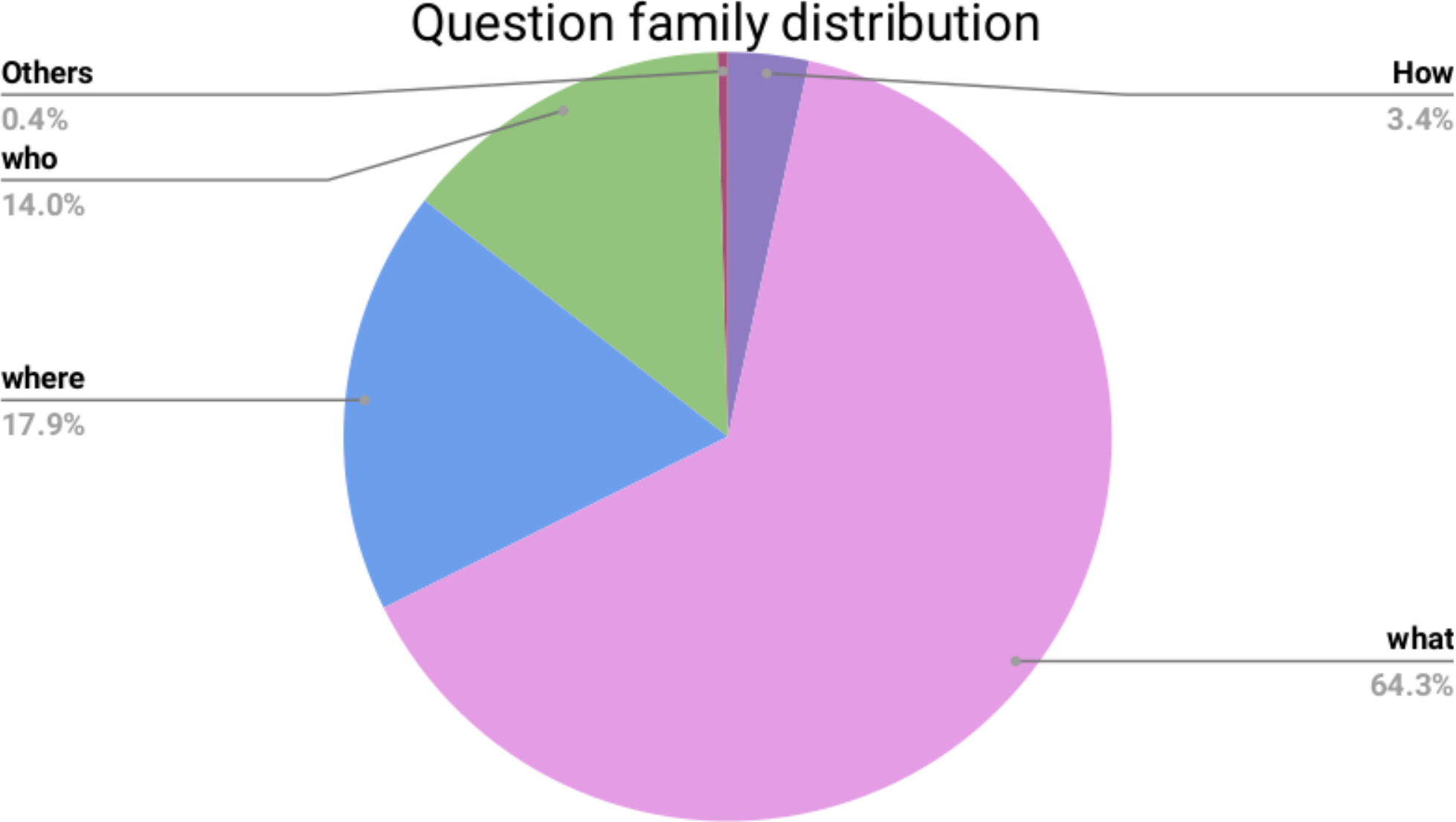}
  \includegraphics[width=0.28\linewidth]{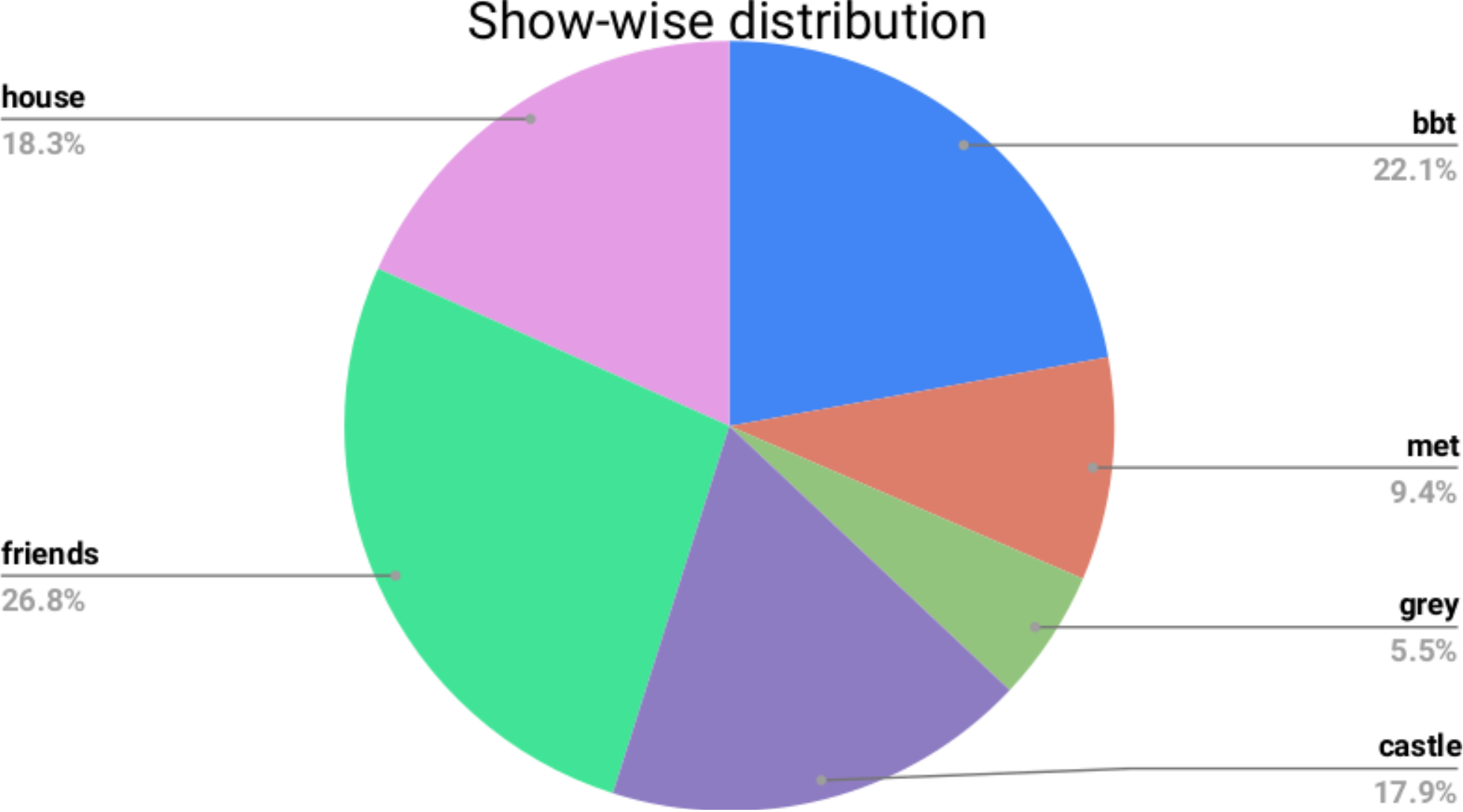}
   \includegraphics[width=0.28\linewidth]{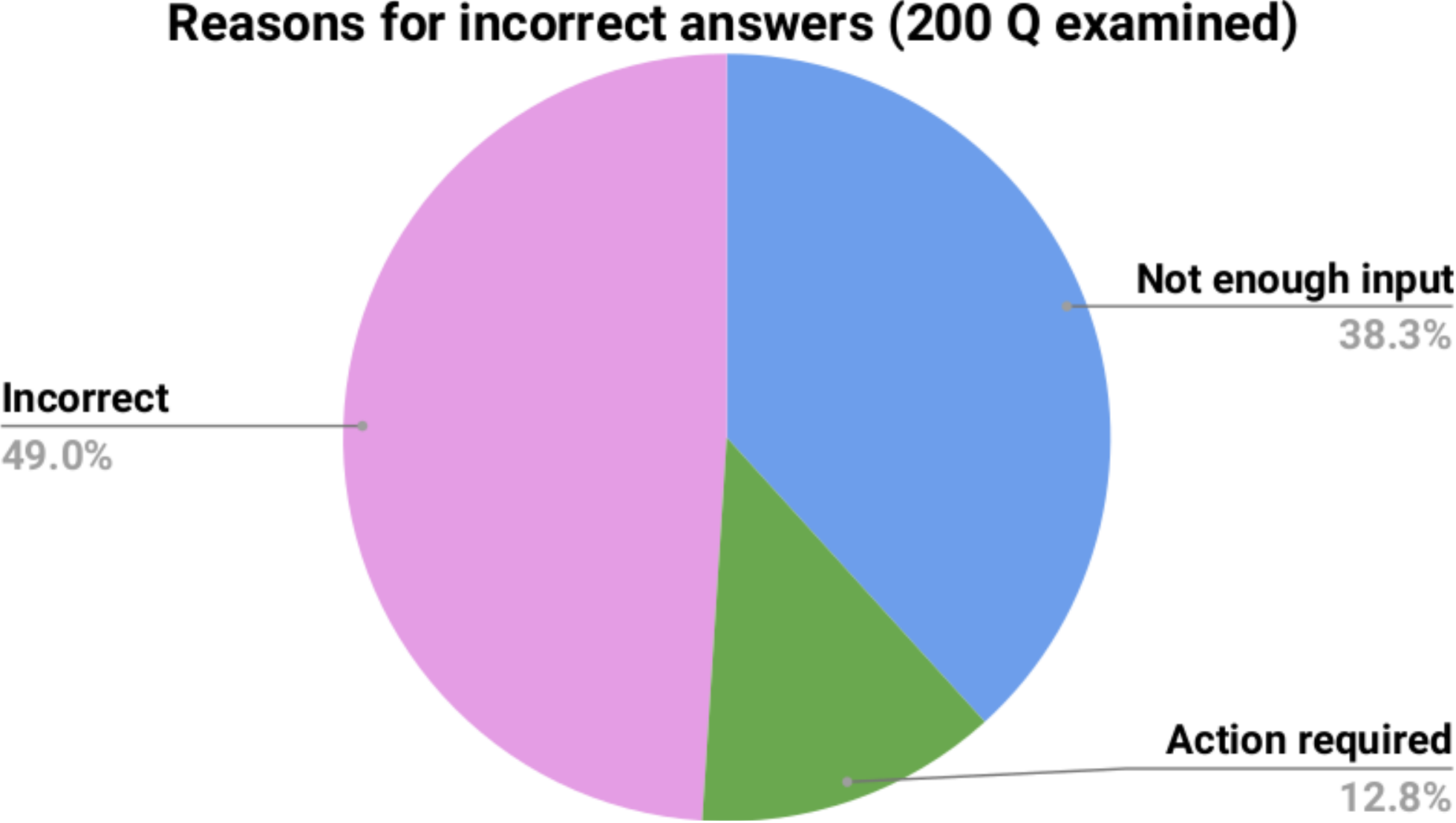}

   \caption{Few statistics for TVQA-Visual (full) set. Left) distribution of questions w.r.t question family, Center) distribution of questions w.r.t TV show, Right) Distribution of questions w.r.t reason of failure. }
\label{fig:visual-set-stats}
\end{figure*}
\input{supp-qual}

\noindent \textbf{Performance analysis on multimodal questions.}
For true multimodal questions which cannot be answered without looking at both video and subtitles, the aggregated feature should rely on both modalities. The last row in figure 5 of the submitted paper is an attempt to study such questions. However, we observed, that many of these type of such questions which apparently require both modalities, can in practice be answered by just one of them. 
Although the question in last row is intended towards both video and dialogue (subtitles), the actual nature of the question is visual. We don't need to know what someone is saying to observe how they are dressed. To the best of our knowledge, no such constraints were imposed while collecting the original TVQA dataset. For instance, a true multimodal question about a specific appearance would be asked if the person appears multiple times with varying appearance in a video. Referring to dialogue in that case to localize the visual input is a true multimodal question. For example, in row 5, the question is " What color is the shirt Marvin is wearing when he say's I could'nt see.?", with the corresponding subtitles, MMFT chooses to ignore subtitles yet giving the correct answer.  For the last row in figure 5, examine the attention maps to see how MMFT gives more attention to V source than subtitles S for the correct answer (which is in the last column).

\noindent \textbf{Fusion Techniques:}
We also tried several other fusion methods including: a) gated fusion where each source vector is gated w.r.t. every other source vectors before fusing them together. We merge the resulting gated source features with i) concatenation followed by a linear layer, ii) taking the product of the gated source vectors, iii) concatenation of the gated fusion feature and the simple fusion feature. All of them result in suboptimal performance than our simple fusion method with a performance drop of 1-2\%. 
\subsubsection{Qualitative Results}
Some of the qualitative results are shown in figure \ref{fig:qualitative} including both success and failure cases of our method and the baselines for Q+V+S input.


\vspace{-5pt}

 



%% file: supp-qual.tex
\begin{figure*}[t]
\footnotesize
\centering
 \includegraphics[width=\linewidth]{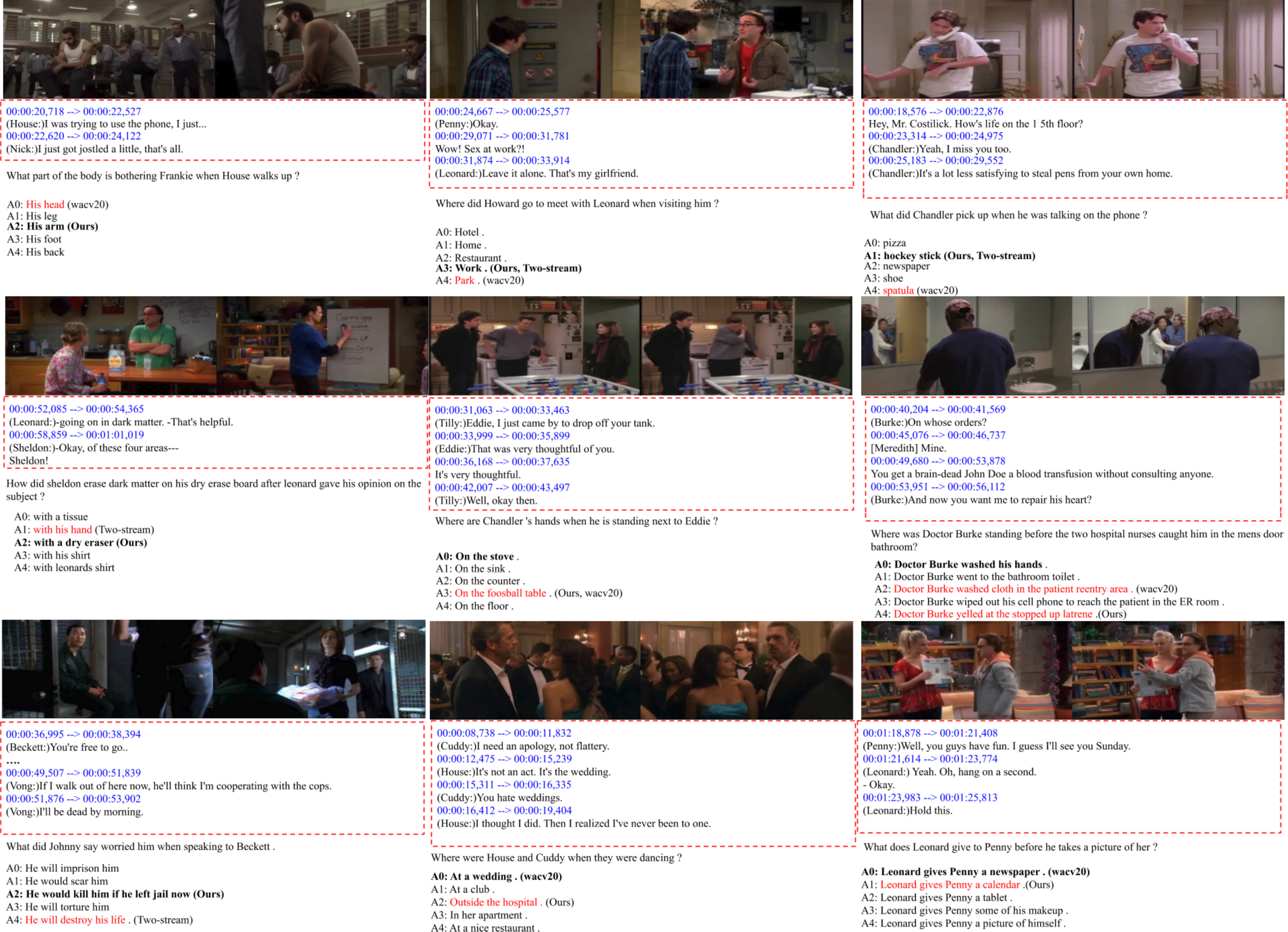}

  \caption{ Qualitative results from validation set. Success and failure cases on visual and multimodal questions.
  Bold text shows correct answer, prediction of each model is in parenthesis. Incorrect prediction is in red font.
  }
   \vspace{-10pt}
\label{fig:qualitative}
\end{figure*}